\documentclass{article}

\usepackage[numbers]{natbib}

    \usepackage[preprint]{neurips_2025}

\usepackage[utf8]{inputenc} 
\usepackage[T1]{fontenc}    
\usepackage{hyperref}       
\usepackage{url}            
\usepackage{booktabs}       
\usepackage{amsfonts}       
\usepackage{nicefrac}       
\usepackage{microtype}      
\usepackage{xcolor}         
\usepackage{amsmath}
\usepackage{graphicx}
\usepackage{arydshln}
\usepackage{makecell}
\usepackage{subcaption}

\title{TRWH: A Text-Driven Random Walk Heterogeneous GNN for Semantic-Aware Sparse Recommendation}

\author{%
  He Ma \\
  The University of Sydney\\
  Camperdown, NSW, 2006, Australia\\
  \texttt{hema5679@uni.sydney.edu.au} \\
  \And
  Chen Liu \\
  Nankai University\\
  Tianjing, China\\
  \texttt{liu.chen@nankai.edu.cn} \\
}

\begin{document}

\maketitle

\begin{abstract}
  Graph Neural Networks (GNNs) and Large Language Models (LLMs) have each advanced recommendation systems by modeling structural and semantic signals, respectively. However, integrating their complementary strengths remains challenging, particularly in sparse settings where maintaining semantic precision is critical. We propose TRWH (Text-driven Random Walk Heterogeneous Graph Neural Network), a novel framework that fuses LLMs-generated textual profiles with heterogeneous graph structures through strategic random walk augmentation. TRWH consists of three core components: (1) Embedding Creation, which produces user and item representations using both Word2Vec and LLMs-based profiling; (2) a Heterogeneous Graph Neural Network (HeteroGNN) that propagates information across multi-relational edges; and (3) Random Walk-based Path Construction, which enriches sparse graphs with second-order user-user and item-item links. Experiments on the Amazon-2023 Fashion (2M users, 825K items) and Beauty (631K users, 112K items) datasets demonstrate that TRWH achieves substantial performance gains over state-of-the-art methods—80.0\% RMSE and 52.6\% MAE reductions on Fashion, and 25.7\% and 10.8\% improvements on Beauty. Notably, while random walks improve performance with traditional embeddings, they can dilute the nuanced representations learned by LLMs—underscoring the importance of adaptive integration strategies.
\end{abstract}

\section{Introduction}
\label{introduction}
Modern recommendation systems face two fundamental challenges: capturing complex user–item relationships in sparse interaction graphs and integrating semantic information from rich textual content. While collaborative filtering effectively mines behavioral patterns, it struggles in cold-start scenarios and fails to leverage the semantic richness of user reviews, item descriptions, and metadata \citep{najafabadi2017improving}. Conversely, content-based methods can utilize textual features but often overlook collaborative signals embedded in interaction patterns.
Graph Neural Networks (GNNs) have emerged as a powerful paradigm for recommendation by modeling user–item interactions as graphs, enabling structured representation learning that captures both direct and higher-order relationships. Their ability to propagate information through graph neighborhoods makes them particularly effective in handling sparse data and cold-start conditions. However, conventional GNNs often operate on interaction graphs alone, missing the wealth of semantic information embedded in unstructured text. In parallel, Large Language Models (LLMs) have demonstrated exceptional capabilities in understanding and generating natural language, enabling new approaches for modeling user preferences and item characteristics from reviews, descriptions, and metadata. Recent work has begun exploring the integration of LLMs into recommendation systems, leveraging their semantic understanding to create textual user and item profiles that reflect nuanced preferences.

Yet, combining GNNs and LLMs introduces several technical challenges: (1) Graph Sparsity: Recommendation graphs are typically sparse. In our dataset, the average number of interactions per user/item is 1.23/3.03 (Fashion) and 1.11/6.23 (Beauty), limiting information propagation effectiveness, (2) Semantic-Structure Integration: Bridging the gap between continuous LLMs embeddings and discrete graph structures, and (3) Computational Efficiency: LLMs-generated features are semantically rich but expensive to process in graph-based learning.

We propose TRWH (Text-driven Random Walk Heterogeneous Graph Neural Network), a novel framework designed to address these challenges through the strategic integration of LLMs-derived semantic features with a multi-relational heterogeneous graph structure. Our contributions are as follows:

(1) Adaptive Semantic Integration: We demonstrate that LLMs-generated profiles can be effectively incorporated into graph neural networks while preserving semantic distinctiveness.

(2) Strategic Random Walk Augmentation: We introduce a one-hop random walk mechanism that enriches sparse graphs by adding second-order user–user and item–item relationships. Notably, we find that such augmentation improves traditional embeddings but may dilute LLMs-derived semantics.

(3) Heterogeneous Graph Architecture: We design a HeteroGNN leveraging diverse edge types (ratings, reviews, purchases, same-store links, and inferred similarities) to model complex user–item interactions.

(4) Comprehensive Empirical Analysis: We conduct extensive experiments on large-scale Amazon 2023 datasets, including ablation studies that reveal the nuanced interactions between embedding types and graph augmentation strategies.

Our experimental results on the Amazon 2023 Fashion and Beauty datasets show that TRWH consistently outperforms state-of-the-art baselines, achieving substantial improvements in RMSE and MAE while maintaining computational efficiency. Furthermore, our analysis reveals that the effectiveness of graph augmentation strategies is contingent on the semantic richness of node embeddings, offering valuable insights for future GNN–LLMs integration in recommendation.
\section{Literature review}
\label{literature}
\subsection{GNNs}
GNNs have revolutionized recommendation systems by modeling user–item interactions as graphs, enabling effective information propagation across structured relationships. Their strength lies in capturing complex information in its structure, which can be harnessed in both homogeneous and heterogeneous graph settings. GNNs have been successfully applied to various graph types, including social networks, item–item graphs, and user–item bipartite graphs, each highlighting different aspects of user behavior and preferences. These graph-based frameworks enable models to learn richer representations and improve recommendation accuracy across a wide range of contexts \citep{feng2025homogeneous, liu2024poi, cui2024rakcr, lonjarret2021sequential, huang2021tag}.

Early methods, such as Neural Collaborative Filtering (NCF) \citep{he2017neural}, laid the groundwork by applying deep learning to collaborative filtering. LightGCN \citep{he2020lightgcn} further advanced the field by simplifying GNN architectures, removing nonlinear transformations and feature projections, and demonstrating that pure neighborhood aggregation could yield state-of-the-art performance.
To better reflect the multi-relational nature of real-world user–item interactions, heterogeneous GNNs have been increasingly adopted. The Heterogeneous Attention Network (HAN) \citep{wang2019heterogeneous} introduced type-specific attention mechanisms, allowing the model to weigh different node and edge types during message passing. Similarly, NGCF \citep{wang2019neural} explicitly captured high-order collaborative signals by propagating embeddings across multiple layers. KGCN \citep{wang2019knowledge} integrated knowledge graphs into GNN architectures, illustrating the benefit of using external relational structures to enhance recommendation accuracy.

Our HeteroGNN design draws architectural inspiration from both HAN and LightGCN. Specifically, we adopt the multi-relational edge modeling strategy from HAN to reflect the diverse nature of user-item interactions(e.g., reviews, purchases, and inferred similarity links). However, instead of using attention across metapaths, we use explicit edge-type modeling for interpretability. From LightGCN, we borrow the principle of simplified message propagation, omitting non-linear transformations to maintain computational efficiency when scaling to millions of nodes.

\subsection{LLMs for recommendation}
Leveraging LLMs for recommendation systems has attracted growing attention \citep{zhao2024recommender, lin2025can, liu2023pre, wu2024survey}, yielding diverse strategies across prompting, profile generation, and embedding integration.
(1) Prompt-based Approaches: Few-shot prompting with ChatGPT \citep{liu2023chatgpt} has shown that large, general-purpose LLMs can perform recommendation tasks without fine-tuning. Chat-REC \citep{gao2303chat} represents user profiles and query contexts as natural language prompts to facilitate explainable and cross-domain recommendation. P5 \citep{geng2022recommendation} introduces a unified text-to-text paradigm that reformulates recommendation tasks—including rating prediction and sequential recommendation—into natural language generation.
(2) Semantic Profile Generation: RLMRec \citep{ren2024representation} utilize LLMs for generating textual profiles of users and items by encoding historical interactions and attributes, an approach that inspired our embedding strategy.
(3) Embedding Integration: Approaches such as UniSRec \citep{hou2022towards} and RecFormer \citep{li2023text} leverage transformer-based architectures to learn universal or task-specific representations directly from interaction and metadata sequences.

Our LLM-based approach in the stage of Embedding Creation is drown from RLMRec. Unlike RLMRec, which generates textual profiles by loosely concatenating past interactions, TRWH adopts a more structured approach by designing system prompts with explicit reasoning instructions to guide the LLMs. Our method applies carefully constructed prompts for both user and item profile generation, incorporating contextual information to ensure semantic richness and relevance. To further enhance representation quality, we employ an instruction-tuned text embedder \citep{su2022one} that transforms these structured profiles into dense vector embeddings, which integrate smoothly with the heterogeneous graph framework. This design not only strengthens semantic alignment with the graph model but also minimizes hallucination risks and enhances the overall coherence and informativeness of the learned representations.
\subsection{Random walk techniques in graph learning}
Random walks have been widely used in graph representation learning, beginning with algorithms such as DeepWalk \citep{perozzi2014deepwalk} and Node2Vec \citep{grover2016node2vec}, which generate random sequences over nodes to learn embeddings via skip-gram models. These techniques preserve local graph structures and community information. In recommendation, Item2Vec \citep{barkan2016item2vec} adapted this principle to model item co-occurrence within sessions. More recent works incorporate random walks into GNNs for better node representation, particularly in sparse or long-tail scenarios. However, these walks are often used for pretraining or embedding generation, rather than augmenting the graph structure during training.

In this paper, we adopt one-hop random walks to introduce semantically proximal links while avoiding noise propagation common in multi-hop walks. Our one-hop random walk strategy introduces second-order proximity in the graph—i.e., indirect user-user and item-item relationships—without significantly increasing graph complexity or introducing semantic drift. The one-hop design offers a balanced trade-off between structural enrichment and semantic stability, especially in sparse graphs where over-propagation of noisy signals can blur fine-grained user or item preferences captured by LLMs.

\begin{figure}[htbp]
  \centering
  \includegraphics[width=1.0\textwidth]{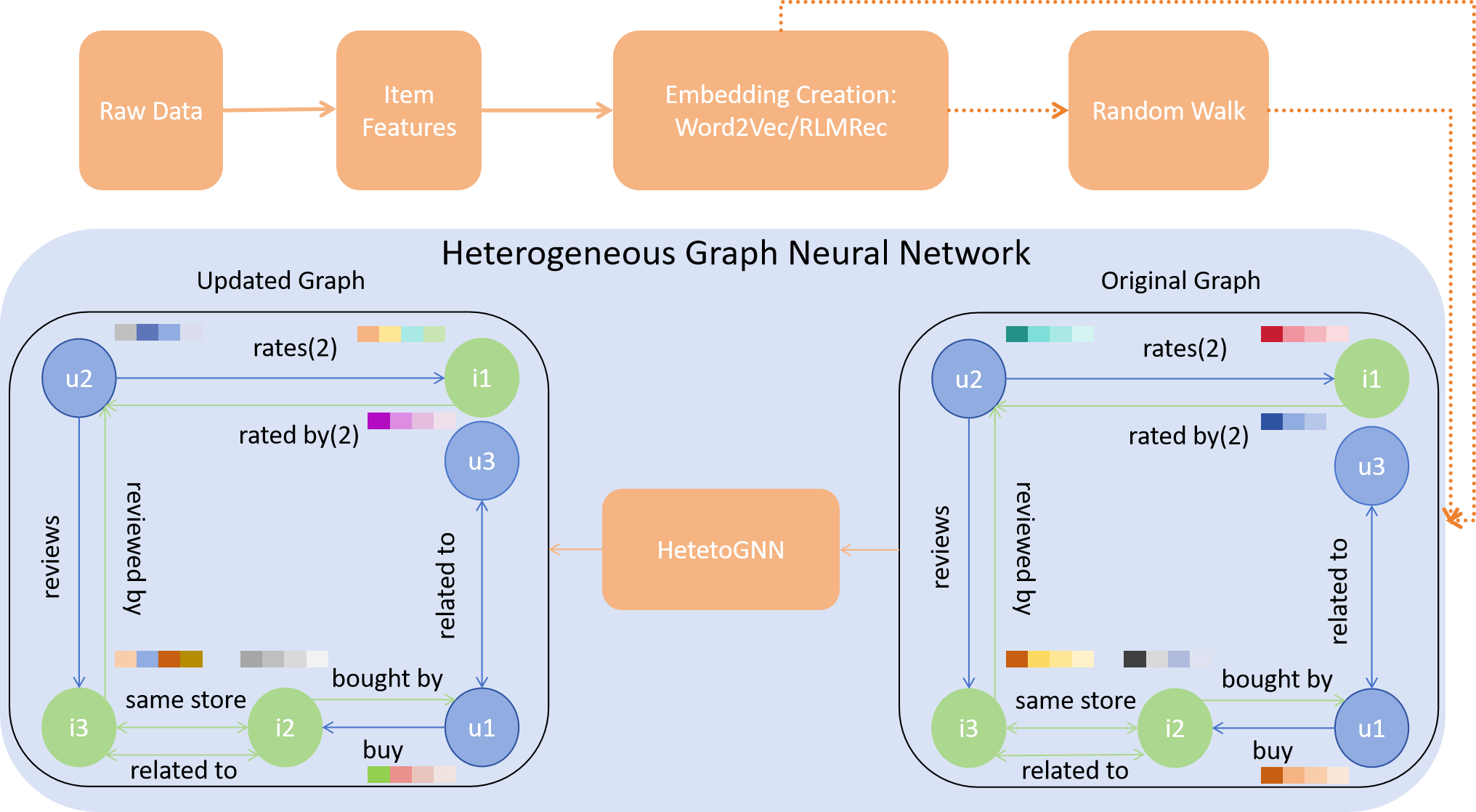}  
  \caption{Text-driven Random Walk Heterogeneous graph neural network(TRWH) main structure}
  \label{fig:mainstructure}  
\end{figure}

\section{Methodology}
The main structure of our proposed methodology-TRWH includes three parts, which are Embedding Creation, Random Walk and Heterogeneous Graph Neural Network respectively. Figure~\ref{fig:mainstructure} shows the working flow of our model. Our model begins with raw data. After data cleaning, relevant item features are selected like average\_rating, average\_number, store, title and etc. These features are used to represent the items in the graph. Next, we feed preprocessed data into Embedding Creation stage to create enriched and meaningful embeddings for users and items. Then, we create a heterogeneous graph with or without random walk. We explain random walk in detail in Section~\ref{randomwalk}. Finally, we train the heterogeneous graph with HeteroGNN. 

\subsection{Embedding creation}
In the Embedding Creation stage, we have tried two different ways to create meaningful embeddings for users and items, which are traditional method-Word2Vec and recent prevalent method LLMs-based approaches. 
\subsubsection{Word2Vec} In this study, we employ the Word2Vec algorithm to capture semantic relationships among words present in both item titles and the review text from users. Specifically, we combine the title of items from the metadata and the review text of users into a unified corpus. Next, combined text is tokenized, resulting in a list of word sequences used to train the Word2Vec model. We configure the model with context window size of 5, minimum word frequency threshold of 1, and utilize 4 worker threads to parallelize the training process. Our finding shows that this setup ensures that even rare words contribute to the embedding space, which aligns with our goal of preserving fine-grained textual signals for downstream tasks.

\begin{figure}[htbp]
  \centering
  \includegraphics[width=1.0\textwidth]{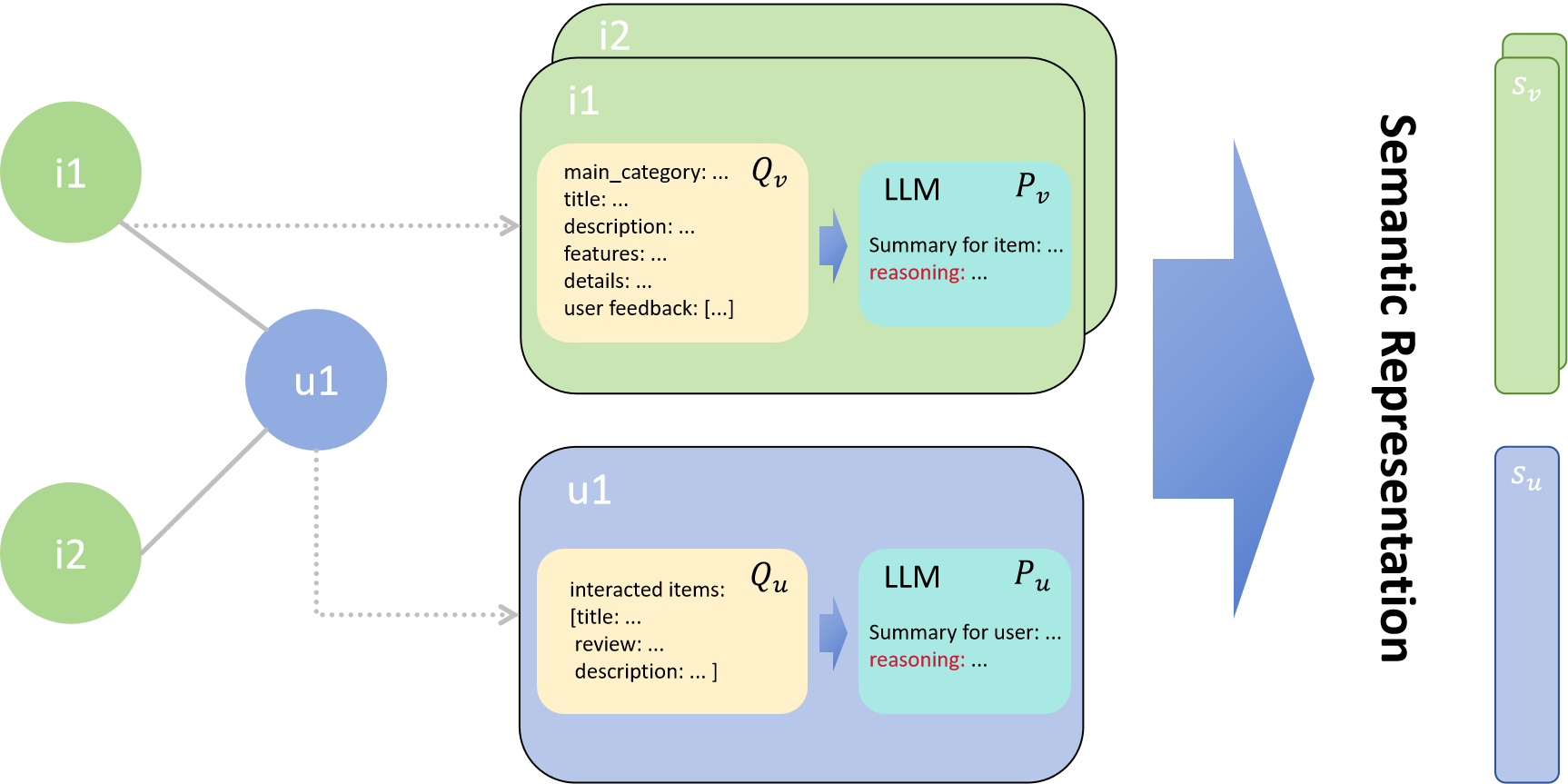}  
  \caption{The method of applying LLM in our model}
  \label{fig:llm}  
\end{figure}

\subsubsection{LLM-based approach} To enhance the expressiveness of item and user representations, inspired by RLMRec \citep{ren2024representation}, we leverage LLMs to generate semantic profiles for users and items.  With the help of LLMs, we obtain useful textual descriptions for users and items. The process of applying LLM is shown in Figure~\ref{fig:llm}. Here we refer these descriptions as profiles for users and items. In this paper, we use Llama-3.2-3B-Instruct to generate profiles for users and items. The final effect of these profiles is as follows: For user profiles, the LLMs encapsulate the user's interacted items that reflect their preferences and their review text about items, enabling the generation of a comprehensive representation of the user’s personalized tastes and interests. For item profiles, it clearly expresses the types of users who are likely to be attracted to the item, offering an insightful depiction of the item’s features and qualities that align with the preferences and inclinations of those users. After generating these profiles, we utilize these profiles to generate corresponding embeddings with a embedder \citep{su2022one}.
\paragraph{Profile generation via reasoning} Recent studies have highlighted the value of integrating reasoning procedures into LLMs to reduce hallucinations and enhance the reliability of generated content. Based on these insights, we carefully designed a system prompt, denoted as \( S_{u/v} \), to be included in the input fed into the LLMs. This prompt is designed to clearly specify the intended role of generating either a user profile for user \(u\) or an item profile for item \(v\), by explicitly outlining the structure of both the input and the expected output. A key aspect of this design is the deliberate incorporation of reasoning steps as a core element of the output. When this system prompt is paired with the corresponding profile generation prompts \(Q_u\) and \(Q_v\), the LLMs can be effectively guided to produce accurate and semantically meaningful profiles. The full procedure is detailed below.
\begin{equation}
\mathcal{P}_u = \mathrm{LLMs}(S_u, Q_u), \quad \mathcal{P}_v = \mathrm{LLMs}(S_v, Q_v)
\end{equation}
\paragraph{Item prompt construction} To construct the input prompt for an item \( v \in \mathcal{V} \), we divide its textual information into six components: the title \( \alpha \), the original description \( \beta \), a set of dataset-specific attributes \( \boldsymbol{\gamma} = \{ \gamma_1, \dots, \gamma_{|\boldsymbol{\gamma}|} \} \), and a set of user reviews \( \mathbf{r} = \{ r_1, \dots, r_n \} \). Based on this categorization, we define the input prompt \( Q_v \) for item profile generation as:
\begin{equation}
Q_v = f_v(\mathbf{x}) \quad \text{w.r.t.} \quad \mathbf{x} = [\alpha, \beta, \boldsymbol{\gamma}, \hat{\mathbf{r}} \subset \mathbf{r}]
\end{equation}
In our method, we define a function \( f_v(\cdot) \), tailored for each item \( v \), which aggregates various textual components into a unified string. By leveraging item attributes information and review text, the resulting prompt provides detailed information to the LLMs, enabling it to generate item profiles that faithfully capture representative and attractive characteristics.
\paragraph{User prompt construction}
We firstly filter out users less than 2 interactions. Next, to construct the profile for a user \( u \), we utilize collaborative signals, we not only utilize the review text of user, but also take advantage of user interacted items-specifically, we select the title and description \( \theta \) of item to generate item profiles. We denote the set of items that user \( u \) has interacted with as \( \mathcal{I}_u \). For each item \( v \in {\mathcal{I}}_u \), we form a textual attribute \( \mathbf{c}_v = [\alpha, \theta, r_u^v] \), where \( r_u^v \) denotes the review written by user \( u \) for item \( v \). The user prompt \( Q_u \) is then formulated as:
\begin{equation}
Q_u = f_u \left( \{ \mathbf{c}_v \mid v \in {\mathcal{I}}_u \} \right).
\end{equation}

The function \( f_u(\cdot) \) serves a similar role as \( f_v(\cdot) \), structuring textual content into a single coherent input. Each component \( \mathbf{c}_v \) includes the user’s review, which reflects their authentic preferences. This composition of the user prompt enables the model to better understand the user's interests. Full details of the prompt design—including \( S \), \( Q \), and \( f_u/f_v(\cdot) \)—as well as illustrative examples, are provided in (create a figure).

\paragraph{Embedding generation}
In this paper, we user an embedder \citep{su2022one} to transform user and item profiles to embeddings. Specifically, we choose instructor-xl\footnote{Link: https://huggingface.co/hkunlp/instructor-xl} as our embedder. We input our generated profiles to embedder, then take their output as embeddings to input latter HeteroGNN nodes. 

\subsection{Heterogeneous graph neural network}
  In this part, we build a heterogeneous graph firstly, then put it into HeteroGNN to train.  The process and structure of our HeteroGNN are shown in Figure~\ref{fig:mainstructure}.

\paragraph{Graph construction} The graph is constructed from the cleaned data, integrating nine distinct edge types. In this heterogeneous structure, blue nodes represent users, while green nodes denote items. "user-rates-item" and "item-rated\_by-user" edges link users to items based on the ratings they have provided. "user-review-item" edges indicate that a user has left a qualitative review for a specific item.  "item-reviewed\_by-user" edges illustrate that a specific item has been left a qualitative review by a user. "user-buy-item" edges signify verified purchases, ensuring that the user has bought and received the item through Amazon. "item-bought\_by-user" Edges link that an item has been bought and received by a user. Since Amazon allows reviews for non-purchased products, "user-buy-item" and "item-bought by-user" edges can help adjust rating reliability. "item-same\_store-item" bidirectional edges connect items from the same store, enabling the model to capture item similarity. "item-related\_to-item" bidirectional edges and "user-related\_to-user" bidirectional edges are bulit by using Random Walk. The concept of Random Walk is covered in Section~\ref{randomwalk}.

\paragraph{Heterogeneous graph neural network training} The constructed graph is fed into a Heterogeneous Graph Neural Network (HeteroGNN), which consists of a single convolutional layer with multiple nodes. During training, nodes within the graph exchange information with their neighbors. For instance, item \(v_1\) shares information with item \(v_2\) through the "same\_store" edge, enabling the HeteroGNN to learn item-level similarities.

\paragraph{Information exchange} After training, nodes in the graph—representing both users and items—have exchanged contextual information. This process enables the model to capture intricate relationships embedded within the data. Such exchanges are reflected visually by changes in item node attributes as shown in the Figure~\ref{fig:mainstructure}. The embedding of each node changes after HeteroGNN training. For example, item \(v\) acquires a more informed representation, shaped by the preferences of user \(u\) and other connected users. Likewise, user \(u\) gains a deeper understanding of the items they have interacted with. This mutual information propagation enhances the model's ability to generate accurate predictions.

\paragraph{Rating prediction} Finally, the trained model outputs the predicted rating, which estimates the score a user is likely to assign to an item. This prediction is based on the learned relationships within the graph.

\subsection{Random walk}
\label{randomwalk}
\begin{figure}[htbp]
  \centering
  \includegraphics[width=0.6\textwidth]{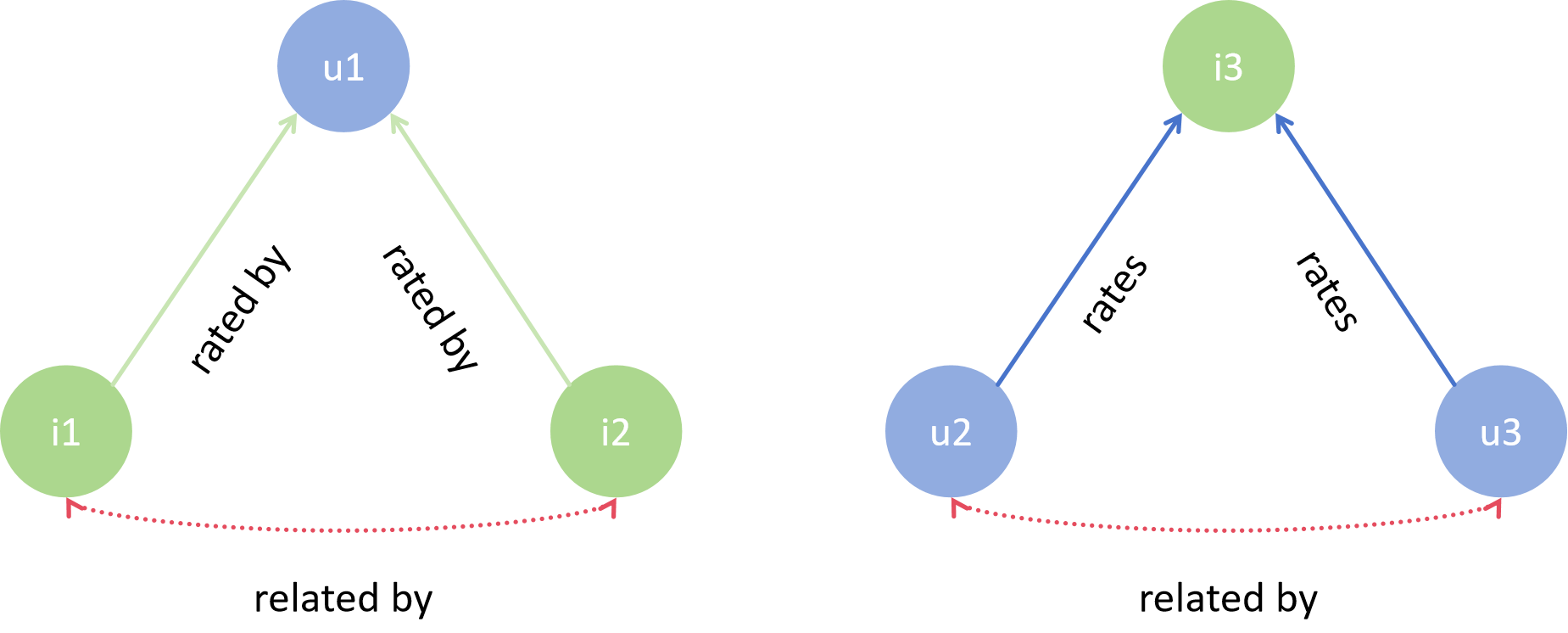}  
  \caption{Random walk figure - it describes how we utilize random walk in our model}
  \label{fig:randomwalk} 
\end{figure}

Our methodology adopts one-hop random walk on two types of edges: "user-rates-item" and "item-rated\_by-user". Figure~\ref{fig:randomwalk} illustrates the core idea behind our one-hop random walk mechanism. In this structure, blue nodes represent users and green nodes represent items, while directed edges reflect interactions between users and items. The left graph of figure shows the process of random walk working in "user-rated\_by-item" edges. The process starts from \( i_1 \), it is rated by plenty of users. We randomly select one user \( u_1 \). Next, because \( u_1 \) is rated by lots of items. We exclude \( i_1 \) firstly, then randomly select one item \( i_2 \). Finally, we create a bidirectional edge called "item-related\_by-item" between \( i_1 \) and \( i_2 \). The right graph of figure illustrates the process of random walk working in "user-rates-item" edges. The process starts from \( u_2 \), it rates plenty of items. We randomly select one item \( i_3 \). Next, because there are lots of other users who rates \( i_3 \). We exclude \( u_2 \) firstly, then randomly select one user \( u_3 \). Then, we build a bidirectional edge called "user-related\_by-user" between \( u_2 \) and \( u_3 \). 

\begin{table}
  \caption{Statistics of Amazon 2023 Fashion and Beauty Dataset}
  \label{tab:amazon2023}
  \centering
  \begin{tabular}{lccccc}
    \toprule
    Dataset & \#Users & \#Items & \#Interactions & \makecell{\#Avg Ratings\\per User/Item} & Rating Distribution\\
    \midrule
    Amazon Fashion  & 2{,}035{,}490 & 825{,}869 & 2{,}500{,}939 & 1.23/3.03 & 1-5 \\
    Amazon Beauty  & 631{,}986 & 112{,}565 & 701{,}528 & 1.11/6.23 & 1-5 \\
    \bottomrule
  \end{tabular}
\end{table}

\section{Evaluation}
\subsection{Evaluation metrics}
This section presents the key metrics used in the study, which are Mean Absolute Error(MAE) and Root Mean Square Error (RMSE) respectively.
\subsubsection{MAE}
MAE quantifies the average magnitude of errors between predicted and actual values. It is computed by measuring the absolute difference between each predicted value \( x_i \) and the corresponding true value \( y_i \). As a widely used metric for regression tasks, MAE reflects the model's prediction accuracy. The objective during model training is to minimize this error, thereby improving prediction reliability.
\begin{equation}
\text{MAE} = \frac{1}{n} \sum_{i=1}^{n} \left| y_i - x_i \right|
\end{equation}
\subsubsection{RMSE}
RMSE is a commonly used metric to evaluate the accuracy of predictive models. It measures the average magnitude of the error by computing the squared differences between the predicted values \( x_i \) and the actual values \( y_i \). These squared errors are then averaged and the square root of that average is taken. This process provides a single value that represents the typical prediction error. The lower the RMSE, the more accurate the model's predictions are. Thus, minimizing RMSE is a key objective in training models.
\begin{equation}
\text{RMSE} = \sqrt{ \frac{1}{n} \sum_{i=1}^{n} (x_i - y_i)^2 }
\end{equation}
\subsection{Results}
Our method---TRWH---includes four different models: Word2Vec+RandomWalk+HeteroGNN (W2VRHet), Word2Vec+HeteroGNN (W2VHet), LLMs+RandomWalk+HeteroGNN (LLMRHet), and LLMs+HeteroGNN (LLMHet).

\subsubsection{Base models}
\paragraph{P\textsuperscript{2}MF}\citep{wang2021cross} is a cross-domain personality-based recommendation system, known as Personality-boosted Probabilistic Matrix Factorization(P$^2$MF). In this framework, Big Five personality scores for users in the target domain are predicted from user reviews using five Support Vector Machines. These predicted traits are then incorporated into the Probabilistic Matrix Factorization (PMF) algorithm to enhance recommendation quality. Specifically, the personality scores are integrated via a multiplicative factor that influences rating behavior, allowing the model to produce more personalized recommendations.

\paragraph{APAR}\citep{yakhchi2020enabling} Analysis of Personality Aspect in Recommender system(APAR) identifies users' personality trait scores by applying the  Linguistic  Inquiry and Word Coun(LIWC) tool to Amazon review texts. The authors introduced a cross-domain recommendation framework that incorporates users’ rating behaviors, personality types, number of reviews, and rating scores. While their method relies on identifying common users across multiple domains to enable domain mapping, they acknowledge that this assumption is often impractical due to the difficulty of finding such overlapping users in real-world scenarios.

\paragraph{PTUPCDR}\citep{zhu2022personalized} Personalized Transfer of User Preferences for Cross-domain Recommendation(PTUPCDR) is a cross-domain recommender system that constructs a meta-network of user embeddings and adopts a task-oriented optimization framework. The model incorporates an attention mechanism to handle both overlapping and non-overlapping users, facilitating the transformation of embeddings from the source domain to the target domain. This design enables PTUPCDR to effectively address cold-start user challenges in the target domain by leveraging transferable user preferences.

\paragraph{CDRIB}\citep{cao2022cross} Cross Domain Recommendation using Information Bottleneck(CDRIB) is a cross-domain recommendation model that transfers knowledge of user–item interactions across domains using information bottleneck (IB) regularizers. The model employs a bipartite graph encoder to capture user–item interactions and generate domain-specific embeddings. By jointly analyzing both cross-domain and intra-domain user–item interactions, as well as user–user correlations, CDRIB is able to learn effective user representations for each domain. Notably, the framework accommodates both overlapping and non-overlapping users across domains to enhance generalization. 

\paragraph{UniCDR}\citep{cao2023towards} This work introduces a unified framework designed to handle diverse cross-domain recommendation (CDR) scenarios by effectively transferring both domain-shared and domain-specific information. The model addresses challenges related to data sparsity and cold-start issues in CDR by supporting overlapping users and items, and by facilitating both intra and inter domain recommendations. UniCDR leverages a contrastive loss function to enhance representation learning across domains, thereby improving recommendation performance in complex cross-domain settings.

\paragraph{NMCDR}\citep{xu2023neural} Node Matching Cross Domain Recommender(NMCDR) presents a framework designed for scenarios involving partially overlapping users across domains. The model integrates intra-domain and inter-domain knowledge through a fully connected user–user graph and utilizes user–item matching information to infer all missing interactions. By combining structural user relationships and interaction data, NMCDR effectively enhances recommendation performance in cross-domain settings.

\paragraph{RealHNS}\citep{ma2023exploring} Real Hard Negative Sampling(RealHNS) introduces a novel model-agnostic framework designed to distinguish false negatives from genuine hard negative samples (HNS)—cases where user–item interactions are missing despite the user having been exposed to the item. The framework leverages negative transfer in cross-domain recommendation scenarios to improve classification accuracy. A dynamic hard negative sampling filter is applied to effectively identify and handle these challenging cases, thereby enhancing the robustness of recommendation models.

\paragraph{MAN}\citep{lin2024mixed} Mixed Attention Network(MAN) introduces an attention-based framework that extracts informative patterns from both same-domain and cross-domain systems. The model employs an encoding layer to capture sequential user behaviors, followed by three types of attention layers to group users across domains. This multi-level attention mechanism enables the prediction of both domain-specific and cross-domain interactions by effectively modeling shared and distinct user preferences.

\paragraph{Homogeneous graph neural network(Homogeneous GNN)}\citep{najafabadi2025theory} This work can predict user ratings for items by utilizing the user–item interaction graph. This process is enhanced through the exchange of information during training. Initially, nodes (users and items) possess only their intrinsic features (e.g., ratings, average\_rating, store). As training progresses, the nodes iteratively update their representations by aggregating information from their neighbors, thereby gaining a richer understanding of their surroundings. This leads to more informed and accurate predictions. Such information propagation is a fundamental advantage of GNNs, as it enables the model to learn complex dependencies within user–item interactions.

\paragraph{ChatGPT (few-shot)}\citep{liu2023chatgpt} This method treats ChatGPT as a general-purpose recommender without any task-specific fine-tuning, leveraging its pretrained knowledge through prompt engineering. 

\paragraph{MF}\citep{koren2009matrix} This paper introduces matrix factorization (MF) as a highly effective technique for collaborative filtering-based recommendation. The key idea is to represent both users and items as vectors in a shared latent factor space, where the interaction between a user and an item is modeled by the dot product of their corresponding vectors. The authors enhance this basic model by incorporating biases (user and item-specific), implicit feedback, temporal dynamics, and varying confidence levels. They propose a regularized optimization objective that minimizes the squared error between observed and predicted ratings. Two optimization strategies are discussed: stochastic gradient descent (SGD) and alternating least squares (ALS), each with its own advantages. The model’s flexibility enables it to integrate various real-world effects, such as changes in user preferences over time and differences in rating reliability, leading to improved prediction accuracy.

\paragraph{MLP}\citep{cheng2016wide} This work is a hybrid recommender system architecture designed to combine the strengths of memorization and generalization. The wide component is a linear model with cross-product feature transformations, effectively memorizing co-occurrence patterns from sparse input features. In contrast, the deep component is a feed-forward neural network that embeds categorical features into low-dimensional dense vectors, allowing the model to generalize to unseen feature combinations. These two components are trained jointly using a shared logistic loss function, enabling the model to learn both low-level feature associations and high-level abstract patterns.

\paragraph{P5}\citep{geng2022recommendation} The framework "Pretrain, Personalized Prompt, and Predict" (P5) introduces a unified text-to-text paradigm that reformulates diverse recommendation tasks—such as rating prediction, sequential recommendation, explanation generation, and direct recommendation—into natural language generation problems. P5 leverages a multitask pretraining strategy using personalized prompts, where user-item interactions and associated metadata are converted into textual inputs and outputs. Built upon an encoder-decoder Transformer architecture, P5 is trained with a shared language modeling objective across tasks, enabling efficient knowledge transfer and zero-shot generalization to unseen users, items, and prompts. This approach significantly enhances the adaptability and scalability of recommendation systems by immersing them in a full language environment and eliminating the need for task-specific architectures.

\paragraph{PEMF-CD}\citep{acharyya2025enhancing} Personality Enabled Matrix Factorization for Cross Domain(PEMF-CD) model is a cross-domain recommendation framework that leverages personality traits inferred from user-generated reviews. It employs a transfer learning strategy to predict personality scores in the target domain using a transformer-based regression model trained on source domain data. Reviews from both domains are jointly embedded via a mixing strategy that clusters similar textual instances to minimize negative transfer. The predicted personality features are then integrated into a probabilistic matrix factorization model, where user neighborhoods are constructed based on a combination of personality-wise and rating-wise similarities. This integration enables the model to effectively handle cold-start scenarios and enhance recommendation accuracy across diverse domains.

\begin{figure}[!htbp]
  \centering
  \begin{subfigure}[t]{0.48\linewidth}
    \centering
    \includegraphics[width=\linewidth]{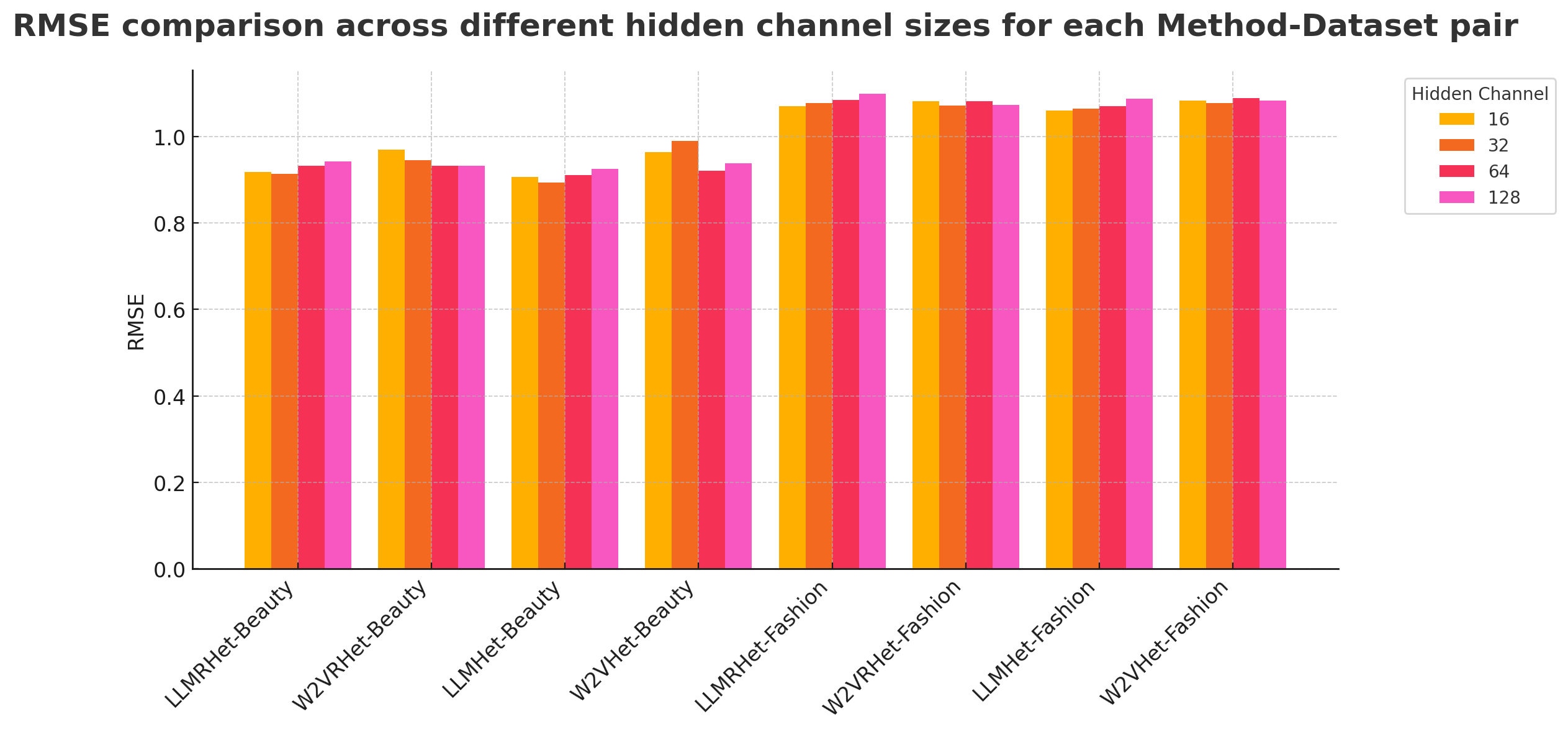}
  \end{subfigure}
  \hfill
  \begin{subfigure}[t]{0.48\linewidth}
    \centering
    \includegraphics[width=\linewidth]{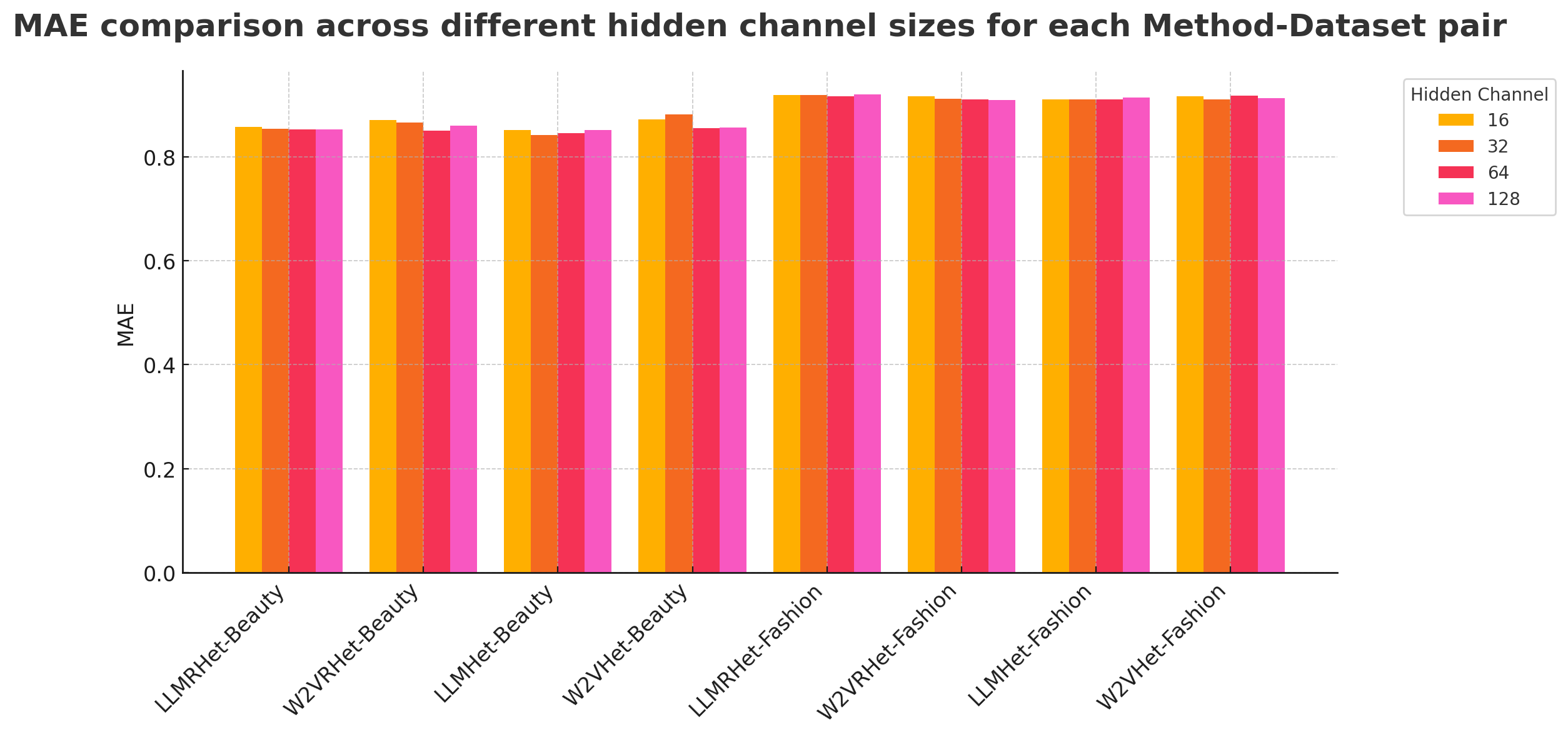}
  \end{subfigure}
  \caption{Comparison of RMSE and MAE across different hidden channel sizes on the Beauty and Fashion datasets}
  \label{fig:rmse-mae}
\end{figure}

\begin{table}[!htbp]
  \centering
  \vspace{\baselineskip}
  \caption{Performance on Fashion and Beauty dataset; [*] denotes our proposed methods}
  \vspace{\baselineskip}
  \label{table:fashion and beauty}
  \begin{minipage}[t]{0.48\linewidth}
    \centering
    \begin{tabular}{lcc}
      \toprule
      \textbf{Model} & \textbf{RMSE} & \textbf{MAE} \\
      \midrule
      P\textsuperscript{2}MF \citep{wang2021cross}       & 11.091 & 2.538 \\
      APAR \citep{yakhchi2020enabling}      & 19.483 & 3.712 \\
      PTUPCDR \citep{zhu2022personalized}         & 7.927 & 1.411 \\
      CDRIB \citep{cao2022cross}  & 7.063 & 1.437 \\
      UniCDR \citep{cao2023towards}  & 7.173 & 1.45 \\
      NMCDR \citep{xu2023neural}    & 8.98 & 3.215 \\
      RealHNS \citep{ma2023exploring}           & 8.64 & 1.520 \\
      MAN \citep{lin2024mixed}   & 10.43 & 5.984 \\
      Homogeneous GNN \citep{najafabadi2025theory} & 1.15   & 1.66 \\
      PEMF-CD \citep{acharyya2025enhancing}      & 6.96 & 1.392 \\
      \midrule
      W2VRHet [*] & 1.0731 & \textbf{0.9089} \\
      LLMRHet [*] & 1.0698 & 0.9191 \\
      W2VHet [*]   & 1.0775 & 0.9105 \\
      LLMHet [*]  & \textbf{1.0604} & 0.9107 \\
      \bottomrule
    \end{tabular}
  \end{minipage}%
  \hfill
  \begin{minipage}[t]{0.48\linewidth}
    \centering
    \begin{tabular}{lcc}
      \toprule
      \textbf{Model} & \textbf{RMSE} & \textbf{MAE} \\
      \midrule
      MF \citep{koren2009matrix}   & 1.1973 & 0.9461 \\
      MLP \citep{cheng2016wide}       & 1.3078 & 0.9597 \\
      P5 \citep{geng2022recommendation}          & 1.2843 & 0.8534 \\
      ChatGPT (few-shot) \citep{liu2023chatgpt}        & 1.0751 & \textbf{0.6977} \\
      Homogeneous GNN \citep{najafabadi2025theory}        & 1.18   & 1.69 \\
      \midrule
      W2VRHet [*]                            & 0.9327 & 0.8496 \\
      LLMRHet [*]   & 0.9134 & 0.8533 \\
      W2VHet [*]  & 0.9204 & 0.8549 \\
      LLMHet [*] & \textbf{0.8944} & \underline{0.8421} \\
      \bottomrule
    \end{tabular}
  \end{minipage}
\end{table}

\vspace{\baselineskip}

\subsubsection{Discussion}
\paragraph{(1) Comparison with baselines:} The results of our four models on both datasets are illustrated in Table~\ref{table:fashion and beauty}. As tables show, on Fashion Dataset, LLMHet and W2VRHet achieves best RMSE and MAE respectively, which significantly outperforms past results. In LLMHet, we utilize LLM-derived profile embeddings encode rich, sentence-level semantics that highlight extreme user-item attitudes. When HeteroGNN propogates these meaningful signals along rating edges, it learns to predict rare but very low (or very high) scores; that suppresses the handful of huge residuals that dominate the squared-error term, so RMSE is lower. At the same time, because the LLM can analyze many subtle details in the text, tiny wording differences can push its predicted rating up or down. That extra error doesn’t ruin any single prediction, but it happens a lot. Mean Absolute Error counts every little slip the same, so all those small misses add up and raise the MAE, so its MAE larger. In W2VRHet, the representations for users and items are shallower: item vectors come from Word2Vec trained only on titles and reviews, while user vectors start as random \texttt{nn.Embedding} parameters. With the help of Random Walk, we add two extra edges, which smooths neighborhood information when messages pass over these extra edges. Smoothing shrinks the absolute number of MAE, hence it obtains a better MAE, which counts every error equally. However, it also blurs the extremes. The model occasionally misses very low or very high ratings, producing a few large squared errors that push RMSE up. On Beauty Dataset, we outperforms in RMSE considerably compared to other models. However, in MAE, our results is higher than ChatGPT(few-shot)~\citep{liu2023chatgpt}. The reason is that our model is trained with MSE loss, we instruct our model to optimize the square of each prediction error. Squaring amplifies outliers: a few large mistakes contribute more to the loss than many small ones. So, our model focuses on shrinking those rare but costly large deviations during training, even if that means tolerating slightly larger absolute errors on the majority of well-predicted cases. Consequently, on the test set, we achieve a lower RMSE which is dominated by large errors and higher MAE which treats all errors uniformly. In contrast, ChatGPT(few-shot) achieves a low MAE because it tends to predict common ratings like 4 or 5 stars, which match the majority of the dataset. This results in small absolute errors for most samples, keeping MAE low. However, since ChatGPT(few-shot) is not trained to minimize squared error, it occasionally makes large mistakes. These rare but severe errors have little effect on MAE but significantly inflate RMSE. \paragraph{(2) Effectiveness of random walk:}As Table~\ref{table:fashion and beauty} illustrates, W2VRHet consistently outperforms W2VHet in most scenarios, with the sole exception being a slightly worse RMSE on the Beauty dataset, which demonstrates that adding more meaningful edges with one hop random walk literally improves the model performance. However, LLMHet outperforms LLMRHet across both metrics and datasets, demonstrating consistently superior performance. Although introducing random walk and adding additional edges typically enriches node representations by aggregating neighborhood information, it can also unintentionally blur critical semantic distinctions embedded by the LLMs and embedder. In LLMHet, node embeddings directly encode precise and nuanced user-item attitudes, maintaining clarity about each node's unique characteristics. When extra edges are introduced through Random Walk in LLMRHet, HeteroGNN begins averaging signals from potentially less-relevant neighbors. This dilutes the distinct semantic signals captured by the LLMs-generated embeddings, reducing the model's sensitivity to extreme or subtle user-item interactions and ultimately causing performance degradation in both RMSE and MAE. \paragraph{(3) Hidden channels:} The results of our four models with different hidden channels are illustrated in Figure~\ref{fig:rmse-mae}. In our experiment, we apply different hidden channels on four models, which are 16, 32, 64 and 128 respectively. The results demonstrate that the effectiveness of hidden channel sizes varies notably across different model configurations and datasets. 
\section{Experiment setup}
\subsection{Dataset}
We evaluate TRWH on two subsets of the Amazon 2023 dataset\footnote{Amazon 2023 dataset Link: https://amazon-reviews-2023.github.io/}. Amazon 2023 dataset is a large-scale Amazon Reviews dataset, collected in 2023 by McAuley Lab, it includes rich features such as: User Reviews, Item Metadata and Links. We select Amazon Fashion and Amazon Beauty dataset to do our experiment. Table~\ref{tab:amazon2023} provides a summary of main components of dataset.

\subsection{Data preparation}
\paragraph{Raw data} Data preparation begins with raw data, it includes all user interactions with items, such as review text. It also has the metadata of items, such as title, categories, average\_rating, average\_number, store and etc.
\paragraph{Preprocessing} We filter irrelevant or useless fields out of data such as (image URLs or fields with a high percentage of null values). Then, we remove user and item whose interactions is less than two on both dataset. Next, key item features are retained by us—specifically, attributes such as average\_rating, average\_number, store, and title from the Amazon dataset. These features are then used to represent items within the graph.

\subsection{Experiment environment and parameters setting}
Our experiments were conducted on both Linux and Windows platforms using Python 3.12. The LLMs-based experiments were conducted on a Linux system equipped with eight 40GB A100 GPUs, while the Word2Vec-based experiments were performed on a Windows system with an RTX 4080 GPU. The CUDA versions used were 12.0 on Linux and 12.6 on Windows. For PyTorch, we used version 2.6.0+cu118 on Linux and 2.7.0+cu126 on Windows. Regarding key hyperparameters, we experiment with learning rates of \{\(1 \times 10^{-3}\), \(1 \times 10^{-4}\), and \(1 \times 10^{-5}\)\} and train for \{\(300\), \(500\), \(800\), \(900\), \(1000\), \(1500\), \(2000\)\} epochs. The number of GNN layers is varied across \{\(1\), \(2\), \(3\)\}. We apply multiple hidden channels which are \{\(16\), \(32\), \(64\), \(128\)\}. We employ 5-fold cross-validation to ensure robust evaluation and apply an early stopping mechanism that halts training when the loss drops below \(0.05\). The Adam optimizer is used for model optimization, and Mean Squared Error (MSE) is adopted as the loss function. Our results are obtained using a learning rate of \(1 \times 10^{-3}\), epochs \(\{900, 1000\}\), GNN layers \(1\), and hidden channels \(\{16, 32, 64, 128\}\). The choice of epoch count and hidden channel size depends on the specific model and dataset used.

\section{Conclusion and Future Work}
\label{conclusion and future work}
In this work, we introduced TRWH, a novel recommendation framework that strategically integrates traditional method-Word2Vec and advanced method-LLMs with HeteroGNN augmented by random walks. TRWH addresses two key challenges in modern recommendation: data sparsity and semantic understanding. Through extensive experiments on the Amazon 2023 Fashion and Beauty datasets, we demonstrate that TRWH significantly outperforms state-of-the-art baselines—achieving up to 80.0\% RMSE and 52.6\% MAE
reductions on Fashion, and 25.7\% and 10.8\% decreases on Beauty. 
\paragraph{Key Insights:}
(1) Embedding-Dependent Augmentation: Random walk augmentation improves traditional embeddings (e.g., Word2Vec) by enriching neighborhood structure, but may dilute the fine-grained semantics captured by LLM-based embeddings.
(2) Heterogeneous Graph Modeling: Incorporating multi-relational edges (rating, review, purchase, same-store, inferred similarity) enhances the model’s ability to learn from diverse interaction signals.
(3) Semantic Profile Integration: LLMs-generated user/item profiles encode nuanced preferences and characteristics often overlooked by collaborative filtering alone, yielding more expressive representations.
\paragraph{Limitations:}
(1) Computational Overhead: LLMs-based profile generation is resource-intensive and may hinder scalability in large-scale production systems.
(2) Scope of Feedback: The current model focuses on explicit feedback (ratings) and does not generalize to implicit signals like clicks or dwell time.
(3) No Cold-Start Evaluation: While sparsity is addressed, user/item cold-start scenarios were not evaluated explicitly.
\paragraph{Future Work:}
(1) Adaptive Random Walks: Design dynamic walk strategies that apply augmentation selectively based on node type, local graph density, and embedding semantics.
(2) Multimodal Integration: Extend the framework to incorporate visual, audio, or structured metadata alongside textual information for richer profiling.
(3) Dynamic Graph Learning: Incorporate temporal information to model evolving user preferences and item popularity over time.
(4) Theoretical Understanding: Develop a theoretical framework to explain when graph augmentation enhances vs. harms semantic embeddings.
(5) Cold-Start Extension: Integrate zero-shot or few-shot methods to improve performance in extreme cold-start settings.

\bibliographystyle{plain}  
\bibliography{reference}

\newpage
\section*{NeurIPS Paper Checklist}

\begin{enumerate}

\item {\bf Claims}
    \item[] Question: Do the main claims made in the abstract and introduction accurately reflect the paper's contributions and scope?
    \item[] Answer: \answerYes{} 
    \item[] Justification: We illustrates our main claims such as methodology and achievements clearly in abstract and introduction, including main parts of methodology and improvement rate on Amazon 2023 Fashion and Beauty dataset, which can accurately demonstrates our contributions and scope.
    \item[] Guidelines:
    \begin{itemize}
        \item The answer NA means that the abstract and introduction do not include the claims made in the paper.
        \item The abstract and/or introduction should clearly state the claims made, including the contributions made in the paper and important assumptions and limitations. A No or NA answer to this question will not be perceived well by the reviewers. 
        \item The claims made should match theoretical and experimental results, and reflect how much the results can be expected to generalize to other settings. 
        \item It is fine to include aspirational goals as motivation as long as it is clear that these goals are not attained by the paper. 
    \end{itemize}

\item {\bf Limitations}
    \item[] Question: Does the paper discuss the limitations of the work performed by the authors?
    \item[] Answer: \answerYes{} 
    \item[] Justification: The limitations of our work are illustrated in Section~\ref{conclusion and future work}.
    \item[] Guidelines:
    \begin{itemize}
        \item The answer NA means that the paper has no limitation while the answer No means that the paper has limitations, but those are not discussed in the paper. 
        \item The authors are encouraged to create a separate "Limitations" section in their paper.
        \item The paper should point out any strong assumptions and how robust the results are to violations of these assumptions (e.g., independence assumptions, noiseless settings, model well-specification, asymptotic approximations only holding locally). The authors should reflect on how these assumptions might be violated in practice and what the implications would be.
        \item The authors should reflect on the scope of the claims made, e.g., if the approach was only tested on a few datasets or with a few runs. In general, empirical results often depend on implicit assumptions, which should be articulated.
        \item The authors should reflect on the factors that influence the performance of the approach. For example, a facial recognition algorithm may perform poorly when image resolution is low or images are taken in low lighting. Or a speech-to-text system might not be used reliably to provide closed captions for online lectures because it fails to handle technical jargon.
        \item The authors should discuss the computational efficiency of the proposed algorithms and how they scale with dataset size.
        \item If applicable, the authors should discuss possible limitations of their approach to address problems of privacy and fairness.
        \item While the authors might fear that complete honesty about limitations might be used by reviewers as grounds for rejection, a worse outcome might be that reviewers discover limitations that aren't acknowledged in the paper. The authors should use their best judgment and recognize that individual actions in favor of transparency play an important role in developing norms that preserve the integrity of the community. Reviewers will be specifically instructed to not penalize honesty concerning limitations.
    \end{itemize}

\item {\bf Theory assumptions and proofs}
    \item[] Question: For each theoretical result, does the paper provide the full set of assumptions and a complete (and correct) proof?
    \item[] Answer: \answerYes{} 
    \item[] Justification: We have provided our assumptions in our Section Methodology. Then, we have discussed our results fully in Section Evaluation, which is complete and correct.
    \item[] Guidelines:
    \begin{itemize}
        \item The answer NA means that the paper does not include theoretical results. 
        \item All the theorems, formulas, and proofs in the paper should be numbered and cross-referenced.
        \item All assumptions should be clearly stated or referenced in the statement of any theorems.
        \item The proofs can either appear in the main paper or the supplemental material, but if they appear in the supplemental material, the authors are encouraged to provide a short proof sketch to provide intuition. 
        \item Inversely, any informal proof provided in the core of the paper should be complemented by formal proofs provided in appendix or supplemental material.
        \item Theorems and Lemmas that the proof relies upon should be properly referenced. 
    \end{itemize}

    \item {\bf Experimental result reproducibility}
    \item[] Question: Does the paper fully disclose all the information needed to reproduce the main experimental results of the paper to the extent that it affects the main claims and/or conclusions of the paper (regardless of whether the code and data are provided or not)?
    \item[] Answer: \answerYes{} 
    \item[] Justification: We already illustrates our key experiment process and parameters in our Section Methodology  and Section Experiment Setup.
    \item[] Guidelines:
    \begin{itemize}
        \item The answer NA means that the paper does not include experiments.
        \item If the paper includes experiments, a No answer to this question will not be perceived well by the reviewers: Making the paper reproducible is important, regardless of whether the code and data are provided or not.
        \item If the contribution is a dataset and/or model, the authors should describe the steps taken to make their results reproducible or verifiable. 
        \item Depending on the contribution, reproducibility can be accomplished in various ways. For example, if the contribution is a novel architecture, describing the architecture fully might suffice, or if the contribution is a specific model and empirical evaluation, it may be necessary to either make it possible for others to replicate the model with the same dataset, or provide access to the model. In general. releasing code and data is often one good way to accomplish this, but reproducibility can also be provided via detailed instructions for how to replicate the results, access to a hosted model (e.g., in the case of a large language model), releasing of a model checkpoint, or other means that are appropriate to the research performed.
        \item While NeurIPS does not require releasing code, the conference does require all submissions to provide some reasonable avenue for reproducibility, which may depend on the nature of the contribution. For example
        \begin{enumerate}
            \item If the contribution is primarily a new algorithm, the paper should make it clear how to reproduce that algorithm.
            \item If the contribution is primarily a new model architecture, the paper should describe the architecture clearly and fully.
            \item If the contribution is a new model (e.g., a large language model), then there should either be a way to access this model for reproducing the results or a way to reproduce the model (e.g., with an open-source dataset or instructions for how to construct the dataset).
            \item We recognize that reproducibility may be tricky in some cases, in which case authors are welcome to describe the particular way they provide for reproducibility. In the case of closed-source models, it may be that access to the model is limited in some way (e.g., to registered users), but it should be possible for other researchers to have some path to reproducing or verifying the results.
        \end{enumerate}
    \end{itemize}

\item {\bf Open access to data and code}
    \item[] Question: Does the paper provide open access to the data and code, with sufficient instructions to faithfully reproduce the main experimental results, as described in supplemental material?
    \item[] Answer: \answerYes{} 
    \item[] Justification: We have already put our dataset access link in Experiment Setup section.
    \item[] Guidelines:
    \begin{itemize}
        \item The answer NA means that paper does not include experiments requiring code.
        \item Please see the NeurIPS code and data submission guidelines (\url{https://nips.cc/public/guides/CodeSubmissionPolicy}) for more details.
        \item While we encourage the release of code and data, we understand that this might not be possible, so “No” is an acceptable answer. Papers cannot be rejected simply for not including code, unless this is central to the contribution (e.g., for a new open-source benchmark).
        \item The instructions should contain the exact command and environment needed to run to reproduce the results. See the NeurIPS code and data submission guidelines (\url{https://nips.cc/public/guides/CodeSubmissionPolicy}) for more details.
        \item The authors should provide instructions on data access and preparation, including how to access the raw data, preprocessed data, intermediate data, and generated data, etc.
        \item The authors should provide scripts to reproduce all experimental results for the new proposed method and baselines. If only a subset of experiments are reproducible, they should state which ones are omitted from the script and why.
        \item At submission time, to preserve anonymity, the authors should release anonymized versions (if applicable).
        \item Providing as much information as possible in supplemental material (appended to the paper) is recommended, but including URLs to data and code is permitted.
    \end{itemize}

\item {\bf Experimental setting/details}
    \item[] Question: Does the paper specify all the training and test details (e.g., data splits, hyperparameters, how they were chosen, type of optimizer, etc.) necessary to understand the results?
    \item[] Answer: \answerYes{} 
    \item[] Justification: We have provided all key parameters and experimental procedures in Section Experiment Setup. Additionally, important details such as the version of the large language model used in our experiments are presented in Section Methodology.
    \item[] Guidelines:
    \begin{itemize}
        \item The answer NA means that the paper does not include experiments.
        \item The experimental setting should be presented in the core of the paper to a level of detail that is necessary to appreciate the results and make sense of them.
        \item The full details can be provided either with the code, in appendix, or as supplemental material.
    \end{itemize}

\item {\bf Experiment statistical significance}
    \item[] Question: Does the paper report error bars suitably and correctly defined or other appropriate information about the statistical significance of the experiments?
    \item[] Answer: \answerYes{} 
    \item[] Justification: We have presented a table and figure to report our RMSE and MAE results, two widely used evaluation metrics in the field of rating prediction.
    \item[] Guidelines:
    \begin{itemize}
        \item The answer NA means that the paper does not include experiments.
        \item The authors should answer "Yes" if the results are accompanied by error bars, confidence intervals, or statistical significance tests, at least for the experiments that support the main claims of the paper.
        \item The factors of variability that the error bars are capturing should be clearly stated (for example, train/test split, initialization, random drawing of some parameter, or overall run with given experimental conditions).
        \item The method for calculating the error bars should be explained (closed form formula, call to a library function, bootstrap, etc.)
        \item The assumptions made should be given (e.g., Normally distributed errors).
        \item It should be clear whether the error bar is the standard deviation or the standard error of the mean.
        \item It is OK to report 1-sigma error bars, but one should state it. The authors should preferably report a 2-sigma error bar than state that they have a 96\% CI, if the hypothesis of Normality of errors is not verified.
        \item For asymmetric distributions, the authors should be careful not to show in tables or figures symmetric error bars that would yield results that are out of range (e.g. negative error rates).
        \item If error bars are reported in tables or plots, The authors should explain in the text how they were calculated and reference the corresponding figures or tables in the text.
    \end{itemize}

\item {\bf Experiments compute resources}
    \item[] Question: For each experiment, does the paper provide sufficient information on the computer resources (type of compute workers, memory, time of execution) needed to reproduce the experiments?
    \item[] Answer: \answerYes{} 
    \item[] Justification: We have provided our experiment environment in Section Experiment Setup.
    \item[] Guidelines:
    \begin{itemize}
        \item The answer NA means that the paper does not include experiments.
        \item The paper should indicate the type of compute workers CPU or GPU, internal cluster, or cloud provider, including relevant memory and storage.
        \item The paper should provide the amount of compute required for each of the individual experimental runs as well as estimate the total compute. 
        \item The paper should disclose whether the full research project required more compute than the experiments reported in the paper (e.g., preliminary or failed experiments that didn't make it into the paper). 
    \end{itemize}
    
\item {\bf Code of ethics}
    \item[] Question: Does the research conducted in the paper conform, in every respect, with the NeurIPS Code of Ethics \url{https://neurips.cc/public/EthicsGuidelines}?
    \item[] Answer: answerYes{} 
    \item[] Justification: We have read NeurIPS Code of Ethics and followed them strictly.
    \item[] Guidelines:
    \begin{itemize}
        \item The answer NA means that the authors have not reviewed the NeurIPS Code of Ethics.
        \item If the authors answer No, they should explain the special circumstances that require a deviation from the Code of Ethics.
        \item The authors should make sure to preserve anonymity (e.g., if there is a special consideration due to laws or regulations in their jurisdiction).
    \end{itemize}

\item {\bf Broader impacts}
    \item[] Question: Does the paper discuss both potential positive societal impacts and negative societal impacts of the work performed?
    \item[] Answer: \answerYes{} 
    \item[] Justification: Our work positively impacts recommendation quality by enhancing semantic understanding and personalization in sparse settings. However, it also poses potential negative societal impacts, such as high computational costs that may lead to increased energy consumption.
    \item[] Guidelines:
    \begin{itemize}
        \item The answer NA means that there is no societal impact of the work performed.
        \item If the authors answer NA or No, they should explain why their work has no societal impact or why the paper does not address societal impact.
        \item Examples of negative societal impacts include potential malicious or unintended uses (e.g., disinformation, generating fake profiles, surveillance), fairness considerations (e.g., deployment of technologies that could make decisions that unfairly impact specific groups), privacy considerations, and security considerations.
        \item The conference expects that many papers will be foundational research and not tied to particular applications, let alone deployments. However, if there is a direct path to any negative applications, the authors should point it out. For example, it is legitimate to point out that an improvement in the quality of generative models could be used to generate deepfakes for disinformation. On the other hand, it is not needed to point out that a generic algorithm for optimizing neural networks could enable people to train models that generate Deepfakes faster.
        \item The authors should consider possible harms that could arise when the technology is being used as intended and functioning correctly, harms that could arise when the technology is being used as intended but gives incorrect results, and harms following from (intentional or unintentional) misuse of the technology.
        \item If there are negative societal impacts, the authors could also discuss possible mitigation strategies (e.g., gated release of models, providing defenses in addition to attacks, mechanisms for monitoring misuse, mechanisms to monitor how a system learns from feedback over time, improving the efficiency and accessibility of ML).
    \end{itemize}
    
\item {\bf Safeguards}
    \item[] Question: Does the paper describe safeguards that have been put in place for responsible release of data or models that have a high risk for misuse (e.g., pretrained language models, image generators, or scraped datasets)?
    \item[] Answer: \answerNA{} 
    \item[] Justification: Our proposed methodology has no such risks.
    \item[] Guidelines:
    \begin{itemize}
        \item The answer NA means that the paper poses no such risks.
        \item Released models that have a high risk for misuse or dual-use should be released with necessary safeguards to allow for controlled use of the model, for example by requiring that users adhere to usage guidelines or restrictions to access the model or implementing safety filters. 
        \item Datasets that have been scraped from the Internet could pose safety risks. The authors should describe how they avoided releasing unsafe images.
        \item We recognize that providing effective safeguards is challenging, and many papers do not require this, but we encourage authors to take this into account and make a best faith effort.
    \end{itemize}

\item {\bf Licenses for existing assets}
    \item[] Question: Are the creators or original owners of assets (e.g., code, data, models), used in the paper, properly credited and are the license and terms of use explicitly mentioned and properly respected?
    \item[] Answer: \answerYes{} 
    \item[] Justification: The existing assets that are used in this paper has been properly credited and respected well.
    \item[] Guidelines:
    \begin{itemize}
        \item The answer NA means that the paper does not use existing assets.
        \item The authors should cite the original paper that produced the code package or dataset.
        \item The authors should state which version of the asset is used and, if possible, include a URL.
        \item The name of the license (e.g., CC-BY 4.0) should be included for each asset.
        \item For scraped data from a particular source (e.g., website), the copyright and terms of service of that source should be provided.
        \item If assets are released, the license, copyright information, and terms of use in the package should be provided. For popular datasets, \url{paperswithcode.com/datasets} has curated licenses for some datasets. Their licensing guide can help determine the license of a dataset.
        \item For existing datasets that are re-packaged, both the original license and the license of the derived asset (if it has changed) should be provided.
        \item If this information is not available online, the authors are encouraged to reach out to the asset's creators.
    \end{itemize}

\item {\bf New assets}
    \item[] Question: Are new assets introduced in the paper well documented and is the documentation provided alongside the assets?
    \item[] Answer: \answerYes{} 
    \item[] Justification: Our paper proposes a new model. We have discussed the details of our model component by component throughout the entire paper. About dataset, we use open source Amazon 2023 Fashion and Beauty dataset. We also provide their link in the paper.
    \item[] Guidelines:
    \begin{itemize}
        \item The answer NA means that the paper does not release new assets.
        \item Researchers should communicate the details of the dataset/code/model as part of their submissions via structured templates. This includes details about training, license, limitations, etc. 
        \item The paper should discuss whether and how consent was obtained from people whose asset is used.
        \item At submission time, remember to anonymize your assets (if applicable). You can either create an anonymized URL or include an anonymized zip file.
    \end{itemize}

\item {\bf Crowdsourcing and research with human subjects}
    \item[] Question: For crowdsourcing experiments and research with human subjects, does the paper include the full text of instructions given to participants and screenshots, if applicable, as well as details about compensation (if any)? 
    \item[] Answer: \answerNA{} 
    \item[] Justification: Our research does not involve crowdsourcing or research with human subjects.
    \item[] Guidelines:
    \begin{itemize}
        \item The answer NA means that the paper does not involve crowdsourcing nor research with human subjects.
        \item Including this information in the supplemental material is fine, but if the main contribution of the paper involves human subjects, then as much detail as possible should be included in the main paper. 
        \item According to the NeurIPS Code of Ethics, workers involved in data collection, curation, or other labor should be paid at least the minimum wage in the country of the data collector. 
    \end{itemize}

\item {\bf Institutional review board (IRB) approvals or equivalent for research with human subjects}
    \item[] Question: Does the paper describe potential risks incurred by study participants, whether such risks were disclosed to the subjects, and whether Institutional Review Board (IRB) approvals (or an equivalent approval/review based on the requirements of your country or institution) were obtained?
    \item[] Answer: \answerNA{} 
    \item[] Justification: Our research does not involve crowdsourcing or research with human subjects.
    \item[] Guidelines:
    \begin{itemize}
        \item The answer NA means that the paper does not involve crowdsourcing nor research with human subjects.
        \item Depending on the country in which research is conducted, IRB approval (or equivalent) may be required for any human subjects research. If you obtained IRB approval, you should clearly state this in the paper. 
        \item We recognize that the procedures for this may vary significantly between institutions and locations, and we expect authors to adhere to the NeurIPS Code of Ethics and the guidelines for their institution. 
        \item For initial submissions, do not include any information that would break anonymity (if applicable), such as the institution conducting the review.
    \end{itemize}

\item {\bf Declaration of LLM usage}
    \item[] Question: Does the paper describe the usage of LLMs if it is an important, original, or non-standard component of the core methods in this research? Note that if the LLM is used only for writing, editing, or formatting purposes and does not impact the core methodology, scientific rigorousness, or originality of the research, declaration is not required.
    \item[] Answer: \answerYes{} 
    \item[] Justification: We have demonstrated our method of using LLM in detail in Section Methodology.
    \item[] Guidelines:
    \begin{itemize}
        \item The answer NA means that the core method development in this research does not involve LLMs as any important, original, or non-standard components.
        \item Please refer to our LLM policy (\url{https://neurips.cc/Conferences/2025/LLM}) for what should or should not be described.
    \end{itemize}

\end{enumerate}

\end{document}